\documentclass[a4paper]{llncs}
\usepackage{amsmath,amsfonts,amssymb}
\usepackage{graphicx}
\usepackage{csvsimple} 
\usepackage{algorithm}
\usepackage{algpseudocode}
\usepackage{theorem}
\usepackage{hyperref}

\newtheorem{de}{Definition}

\newcommand{\ie}{\textit{i.e.~}}

\newcommand{\al}{\textit{al.~}}
\newcommand{\espace}{\vspace{10pt}}

\begin{document}

\title{Mining relevant interval rules}

\author{Thomas Guyet\inst{1} \and Ren\'e Quiniou\inst{2} \and V\'eronique Masson\inst{3}}

\institute{AGROCAMPUS-OUEST/IRISA-UMR 6074
\and
Inria, Centre de Rennes
\and
University Rennes-1/IRISA-UMR 6074}

\maketitle

\begin{abstract}
This article extends the method of Garriga et \al for mining relevant rules to numerical attributes by extracting interval-based pattern rules. We propose an algorithm that extracts such rules from numerical datasets using the interval-pattern approach from Kaytoue et \emph{al.}
This algorithm has been implemented and evaluated on real datasets.
\end{abstract}

\section{Introduction}
Garriga et \al \cite{Garriga2008} proposed a method to extract relevant association rules from labeled itemsets. 
We extend the work of Garriga et \al to numerical attributes using the pattern mining approach of Kaytoue et \al \cite{Kaytoue2011} which is based on FCA (Formal Concept Analysis). Kaytoue et \al \cite{Kaytoue2011} proposed to extend the mining of frequent closed interval pattern to numerical data. Our work bridges the gap between these two approaches to extract relevant interval pattern rules.

\section{Closed interval patterns}
Let $\mathcal{F} = \{f_1 , \dots , f_n \}$ be a fixed set of $n$ features. We represent a training example as a tuple of real values $\vec x=\{x_1, \dots, x_n\}$, where $x_i \in Dom(f_i)$, with an associated class label. The tuple stores one value per feature of $\mathcal{F}$. 
We consider two-class learning problems where the set of examples $E$ is divided in positives ($P$) and negatives ($N$) such that $E = P \cup N$ and $P \cap N = \emptyset$. 
Multi-class problems can be transformed in two-class learning problems.

An $n$-dimensional interval pattern is a tuple of intervals $\langle[l_i, u_i]\rangle_{i\in[1,...n]}$, where $l_i, u_i \in \mathcal{M}_i\subset\mathbb{R}$, $l_i \leq u_i$ and $\mathcal{M}_i$ is an ordered finite set of modalities (\ie each $\mathcal{M}_i$ is a set of feature values, \ie $Dom(f_i)$, or a subset of values $\mathcal{M}_i\subset Dom(f_i)$). 
An interval pattern $P=\langle [l_i, u_i]\rangle_{i\in[1,...n]}$ covers a tuple $\vec{x}=\{x_1, \dots, x_n\}$, denoted $ \vec x \sqsubseteq P$, iff $\forall i\in[1,...n],\; l_i < x_i \leq u_i$.

Let $X=\langle[l_i, u_i]\rangle_{i\in[1,...n]}$ and $Y=\langle[l'_i, u'_i]\rangle_{i\in[1,...n]}$ be two $n$-dimensional interval patterns. We define $X\sqcup Y = \left\langle \left[\min\left(l_i, l'_i\right), \max\left(u_i, u'_i\right)\right]\right\rangle_{i\in[1,...n]}$. Further, $X\sqsubseteq Y$ iff $\forall i\in[1,...n],\; [l_i, u_i] \subseteq [l'_i, u'_i]$. This definition extends the previous one for tuple covering considering a value $v$ as a singleton interval $[v,v]$.

Let $X\rightarrow +$ be a positive rule where $X$ is an interval pattern. 
True positives are positive examples covered by the rule: $TP(X)=\{\vec e|\vec e\in P \wedge \vec e \sqsubseteq X\}$. False positives are negative examples covered by the rule: $FP(X)=\{\vec e|\vec e\in N \wedge \vec e\sqsubseteq X\}$. True negatives are negative examples not covered by the rule: $TN(X)=\{\vec e|\vec e\in N \wedge \vec e\not\sqsubseteq X\}$.
$supp(X)$, the support of pattern $X$ is defined as $supp(X)=|\{\vec e|\vec e\in E \wedge \vec e\sqsubseteq X\}|$. We also define $supp^+(X) = |TP(X)|$ and $supp^-(X) = |FP(X)|$. 
$supp^+$ is antimonotone w.r.t the $\sqsubseteq$ relation and $supp^-$ is monotone w.r.t $\sqsubseteq$. This means that $\forall X,Y,\; X\sqsubseteq Y,\; supp^+(X)\leq supp^+(Y)$ and $supp^-(Y)\leq supp^-(X)$.

The learning task consists in constructing all interval patterns $X$ such that $supp^+(X)>minsup$ and $supp^-(X)<maxfp$ where $minsup$ and $maxfp$ are given parameters. 


From the practical point of view of data mining algorithms, closed patterns are the largest patterns (w.r.t. a partial order $\sqsubseteq$ on the set of patterns, denoted $\mathcal{P}$) among patterns occurring in the exact same set of examples. Formally, 
a set $X\in\mathcal{P}$ is closed when there is no other set $Y\in\mathcal{P}$ such that $X \sqsubset Y$ (\ie $Y \sqsubseteq X \wedge Y \neq X$) and $supp(X) = supp(Y)$. Closed patterns are interesting because they carry the same information as the total set of frequent patterns.

Kaytoue et \al \cite{Kaytoue2011} have investigated the problem of mining frequent closed interval patterns with Formal Concept Analysis (FCA). They proposed the \Call{MinIntChange}{} algorithm which enumerates all frequent closed frequent patterns. It starts from the most generic interval pattern that covers all the examples: $IP = \left\langle \left[\min\left(\mathcal{M}_i\right), max\left(\mathcal{M}_i\right)\right]\right\rangle_{i\in 1 \dots n}$. Then, each interval pattern is specialized applying minimal changes on the left or on the right of the interval.

\section{Mining relevant interval-rules}
The theory of relevancy, described in \cite{Lavrac2005}, aims mainly at reducing the hypothesis space by eliminating irrelevant features. This theory has been used by Garriga et \al \cite{Garriga2008} to extract relevant features in example database where an example is a tuple of symbolic features.
Here, we extend the definition of relevancy of Garriga et \al \cite{Garriga2008} to the relevancy of interval patterns. First, we define two closure operators, $\Gamma^+$ and $\Gamma^-$, that respectively stand for the closure of interval pattern on $P$ (positive examples) and on $N$ (negative examples).

\begin{de}[Relevancy of an interval pattern]\label{def:relevancy}
Let $X$ and $Y$ be two interval patterns. $X$ is more relevant than $Y$ iff $\Gamma^+\left(Y\right)=\Gamma^+\left(X\sqcup Y\right)$ and  $\Gamma^-\left(X\right)=\Gamma^-\left(X\sqcup Y\right)$.
\end{de}

Thus, similar results as those of Garriga et \al \cite{Garriga2008} can be deduced about the characterization of the space of relevant interval patterns.

\begin{theorem}\label{th:positiveclosed}
Let $X$ and $Y$ be two interval patterns. If $\Gamma^+(Y)=X$ and $Y\neq X$ then $Y$ is less relevant than $X$.
\end{theorem}

\begin{theorem}\label{th:negative}
Let $X$ and $Y$ be two different closed interval patterns such that $X \sqsubset Y$. Then, we have that $Y$ is less relevant than $X$ iff $\Gamma^-(X)=\Gamma^-(Y)$.
\end{theorem}

The first theorem shows that the relevant rules $X \rightarrow +$ are those for which the interval pattern $X$ is closed over the positive examples. According to the second theorem, in case of similar negative supports, the interval pattern with largest intervals is preferred.
Proofs for Theorems \ref{th:positiveclosed} and \ref{th:negative} may be deduced from proofs on features sets \cite{Garriga2008}.

Algorithm \ref{alg:closedintervalrule} is based on these theorems to extract the relevant interval patterns. The first step of the algorithm is to extract $FCIP$, the set of frequent interval patterns closed over the positives. 
Then, line 3 prunes irrelevant patterns in accordance with Theorem \ref{th:negative}. For any closed interval pattern $Y \in FCIP$, if there exists another closed interval pattern $X$ such that both have the same support in the negatives (\ie same number false-positives) and such that $X \sqsubset Y$ then $Y$ is removed.

\begin{algorithm}[tb]
\caption{Closed interval rule mining algorithm. $P$ is the set of positive examples, $N$ is a set of negative examples and $\mathcal{M}$ is the set of modalities.}\label{alg:closedintervalrule}
\begin{algorithmic}[1]
\State $FCIP \gets$\Call{MinIntChange}{$P$, $\mathcal{M}$}
\For{$(X, Y) \in FCIP$}
\If{$FP(X)=FP(Y)$ \textbf{and} $X\sqsubset Y$}
\State $FCIP \gets FCIP\setminus\{Y\}$
\EndIf
\EndFor
\end{algorithmic}
\end{algorithm}

\espace

The size of the interval patterns search space is $O\left(m^{2\times n}\right)$ where $n$ is the number of features and $m$ is the number of modalities $\mathcal{M}_i$ of one attribute. Thus, we are facing a memory usage constraint. Keeping all the frequent concept in memory require a large memory. This memory issue is classically encountered in formal concept analysis but it becomes harder when the number of modalities increases.

To tackle the issue of memory usage, we reduce the modalities to a subset $\mathcal{M}_i$ of a fixed maximal size, defined by parameter \textit{eqmod}. The overall rule mining algorithm has not to be modified. There are several methods to reduce the number of modalities. We choose to extract the equi-probable intervals from the positives examples.

\section{Implementation and results}
We evaluated our algorithm on three UCI datasets \cite{UCI2013} (\textit{Haberman}, \textit{Iris} and \textit{Vertebral column}). The algorithm is implemented in C++. Experiments are conducted on an Intel Core-I5 with 8Go of RAM with Linux system.

\begin{figure}[tb]
\centering
\includegraphics[width=0.31\textwidth]{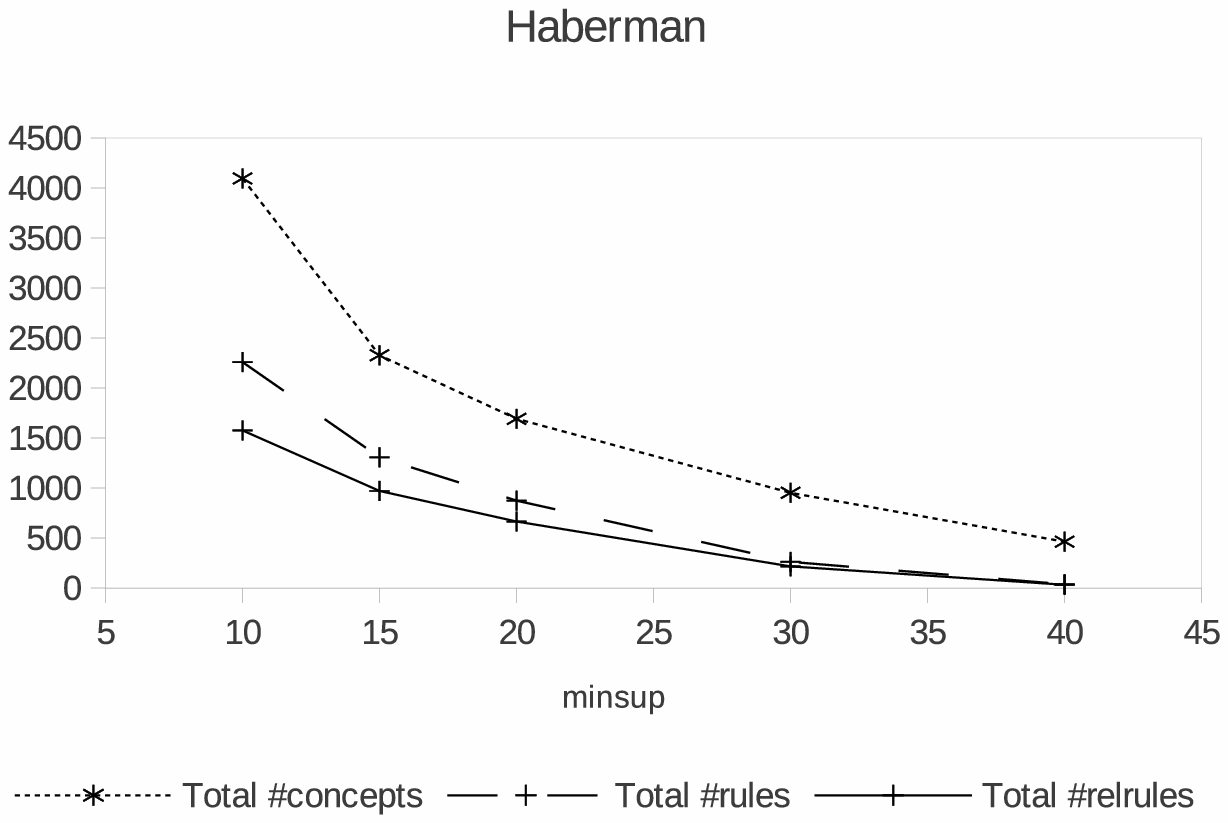}
\includegraphics[width=0.31\textwidth]{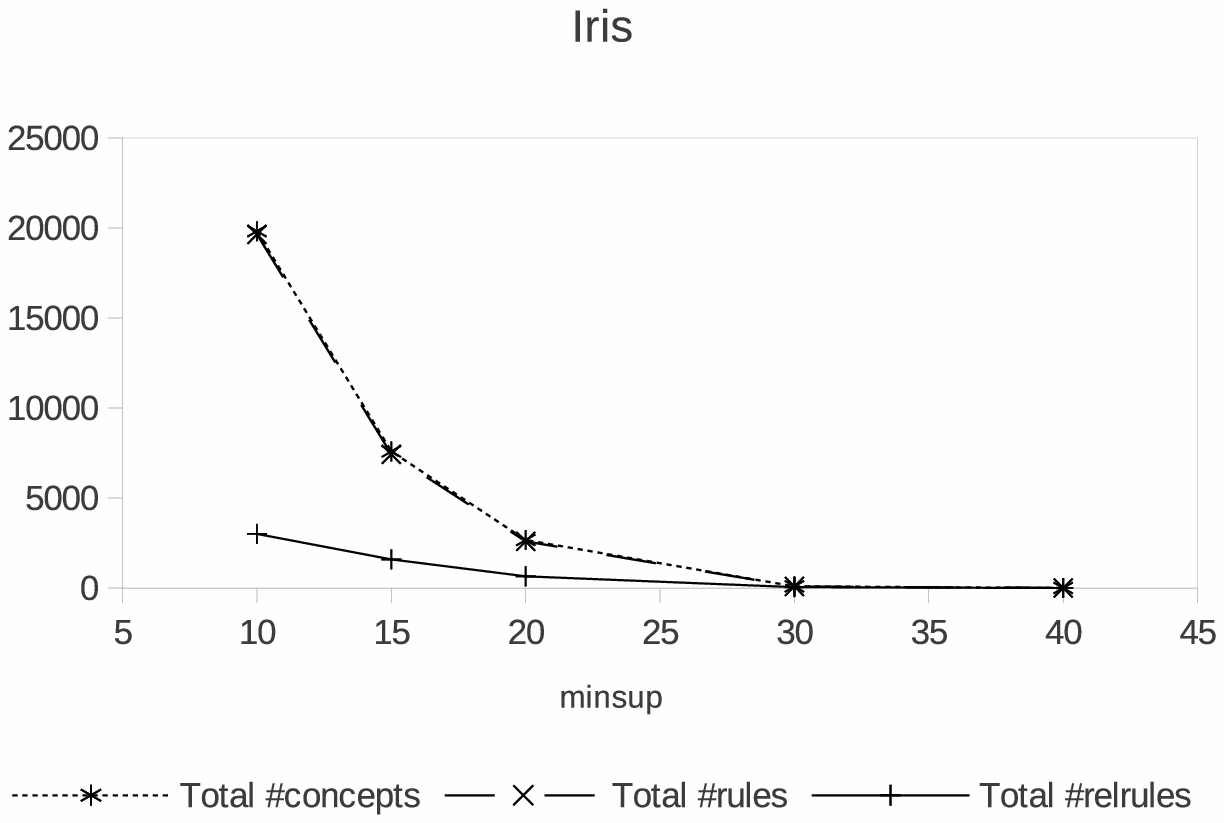}
\includegraphics[width=0.31\textwidth]{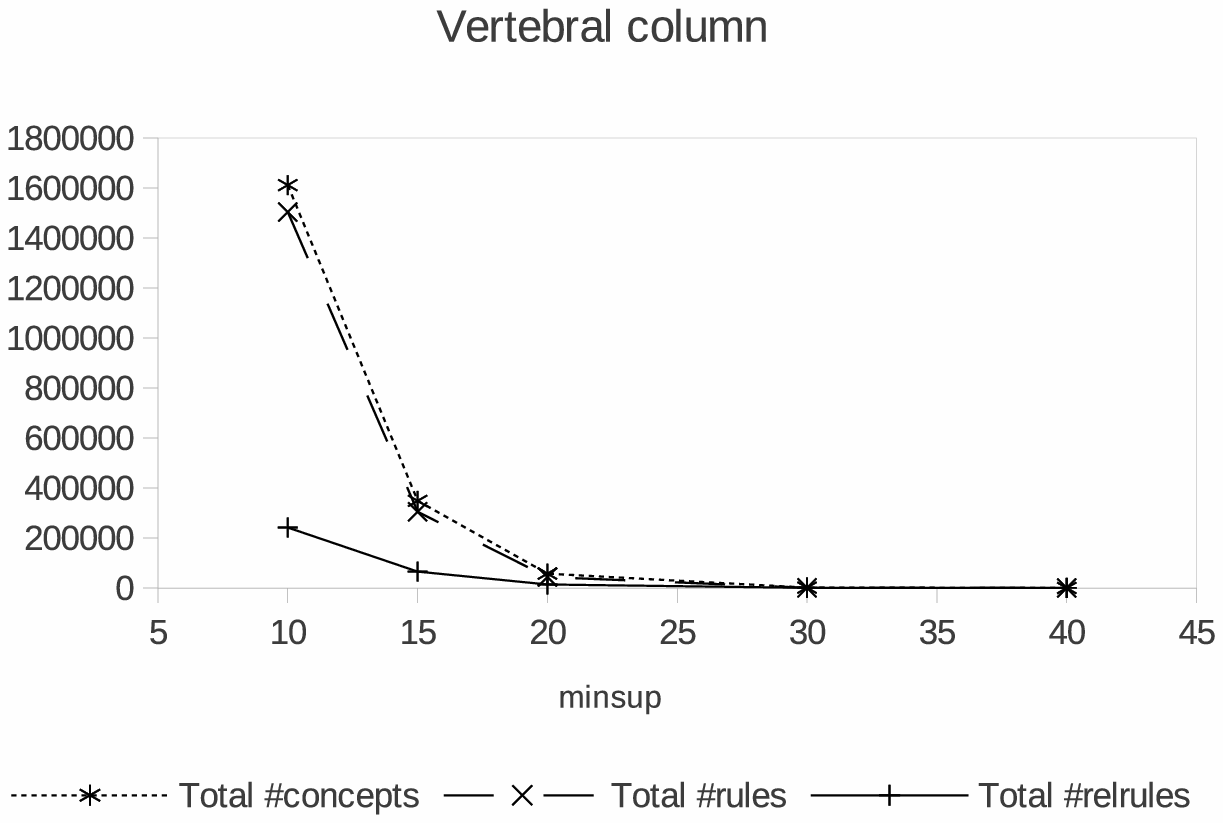}
\caption{Number of closed interval patterns in positives, number of rules satisfying $minsup$ and $maxfp$; and number of relevant interval-rules w.r.t. minimal support.}
\label{fig:NbMinsup}
\end{figure}

\begin{figure}[tb]
\centering
\includegraphics[width=0.31\textwidth]{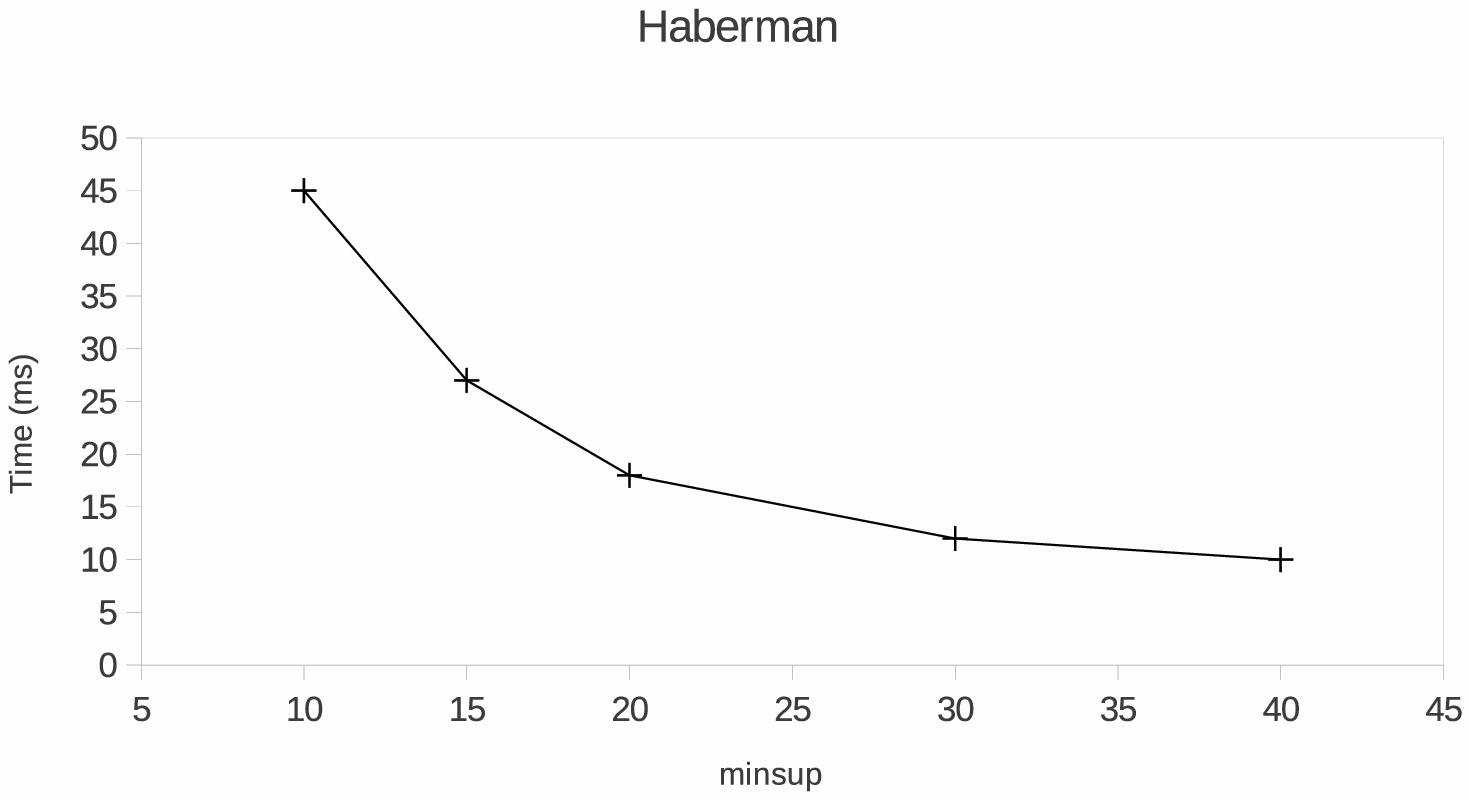}
\includegraphics[width=0.31\textwidth]{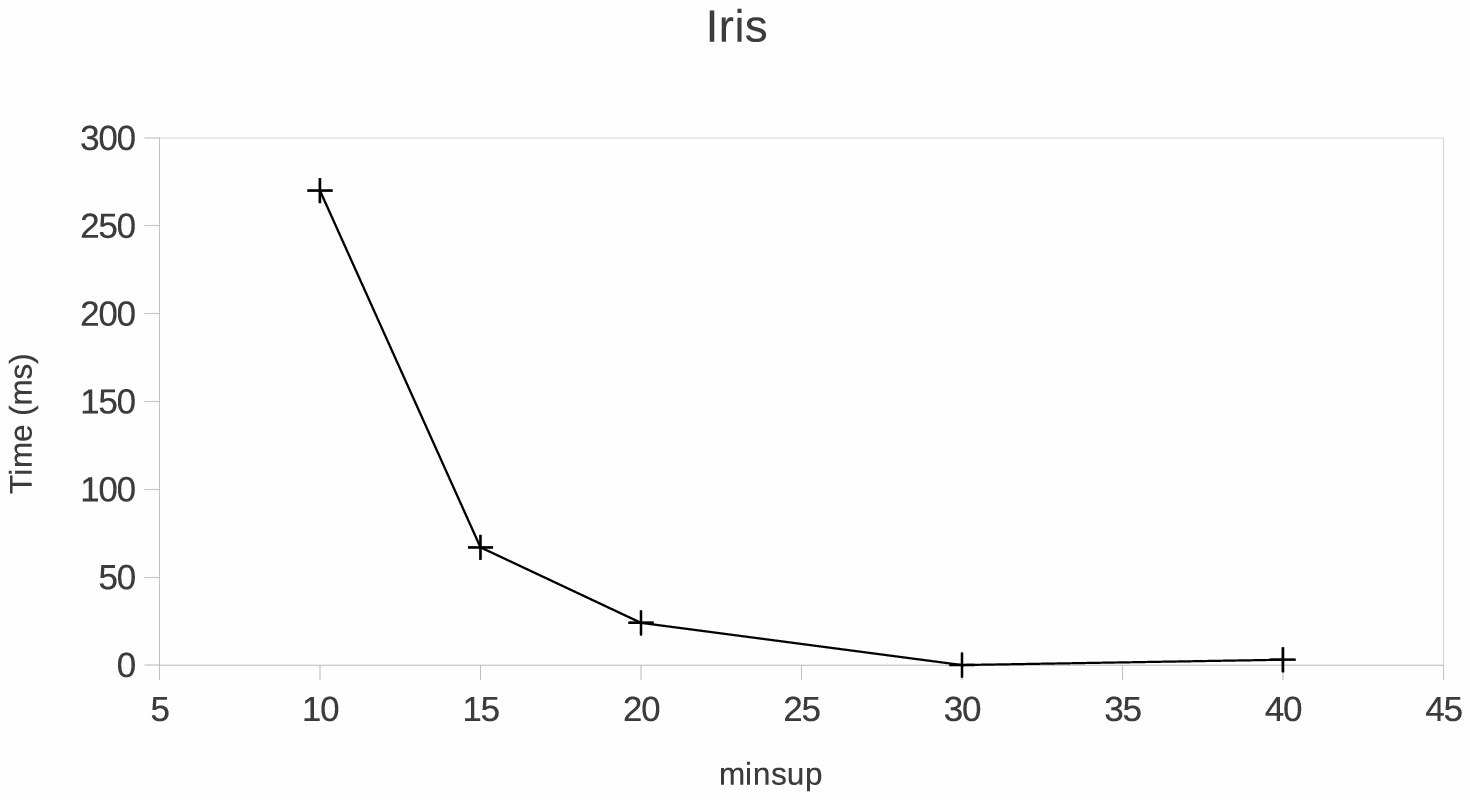}
\includegraphics[width=0.31\textwidth]{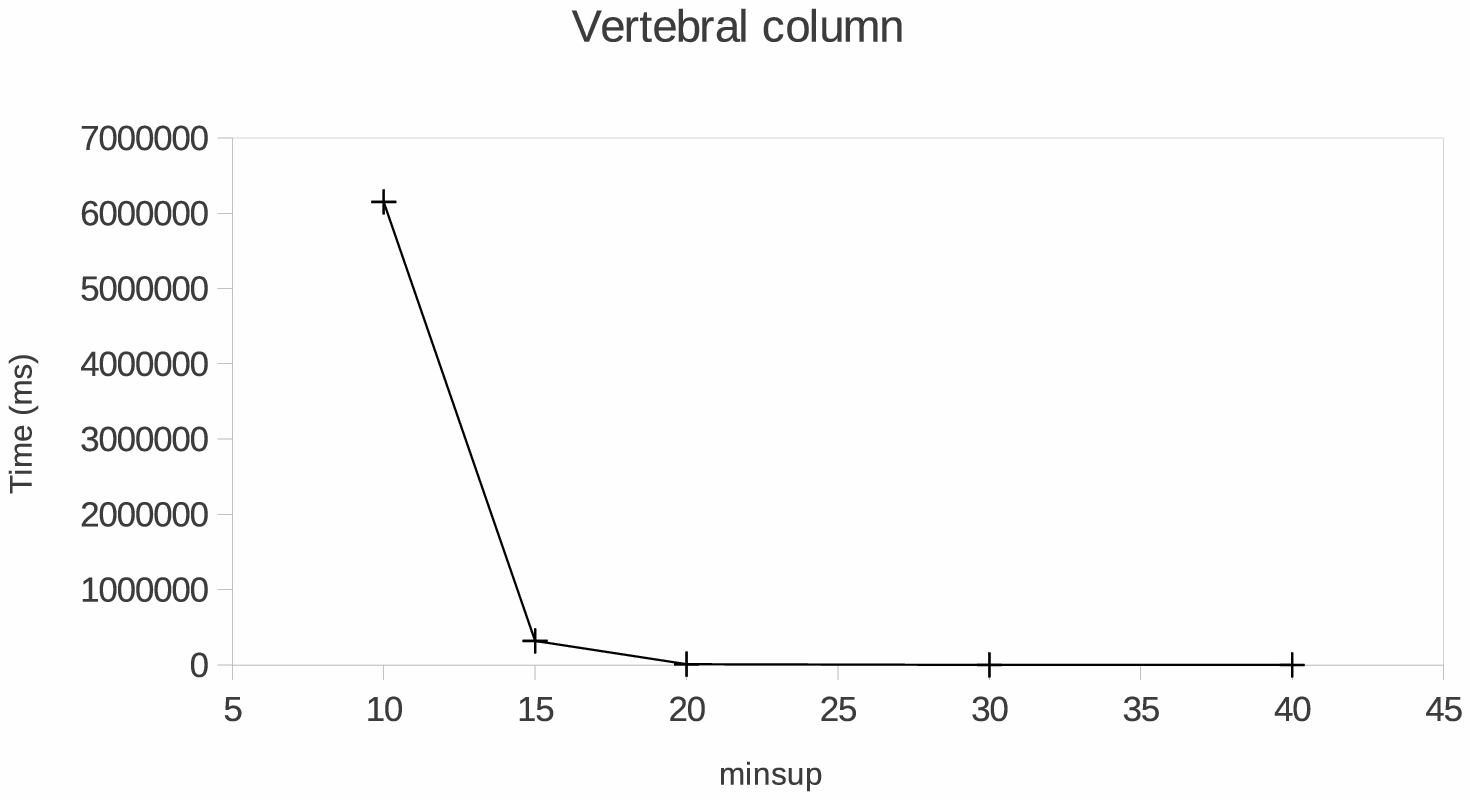}
\caption{Computing time (in millisecond) w.r.t. minimal support.}
\label{fig:TimeMinsup}
\end{figure}

For all experiments in this section, $fpmax=10\%$ and $eqmod=10$. 
Figure~\ref{fig:NbMinsup} illustrates the number of closed interval patterns in positive examples, the number of frequent and accurate rules; and the number of relevant interval pattern rules. 
We can see that the computing times (see Figure~\ref{fig:TimeMinsup}) are strongly correlated to the number of patterns.

Even for small data such as the Iris dataset, the number of patterns is high for low thresholds ($\approx 3000$) but the number of relevant patterns is significantly lower than the total number of closed rules. 
Moreover, the number of patterns increases exponentially with the number of modalities.

\section{Conclusions}
We have presented a new algorithm for extracting relevant rules from a numerical dataset. It offers a wider choice of possibly interesting rules for experts. The number of extracted patterns is high but more representative of the input dataset whereas standard algorithms such as CN2 or Ripper select a priori a very limited set of rules simply based on covering and accuracy criteria. 
Future work will be devoted to proposing additional selection criteria which enable the expert to express his/her preferred set of relevant rules.

\bibliographystyle{plain}
\bibliography{biblio}

\end{document}